%% file: con_mdsgd.tex
\documentclass[twoside,11pt]{article}

\usepackage{jmlr}
\usepackage{amsmath}
\usepackage{algorithmic}
\usepackage{algorithm}
\usepackage{mathrsfs}
\usepackage{booktabs}       
\usepackage{amsfonts}       
\usepackage{nicefrac}       
\usepackage{microtype}      
\usepackage{subfigure}

\jmlrheading{1}{2000}{1-48}{4/00}{10/00}{meila00a}{Shen-Yi Zhao, Hao Gao and Wu-Jun Li}

\include{defs}

\begin{document}

\title{On the Convergence of Memory-Based Distributed SGD}

\author{\name Shen-Yi Zhao \email zhaosy@lamda.nju.edu.cn \\
        \name Hao Gao \email gaoh@lamda.nju.edu.cn \\
        \name Wu-Jun Li \email liwujun@nju.edu.cn \\
        \addr Department of Computer Science and Technology \\
              Nanjing University, China}


\maketitle

\begin{abstract}
  Distributed stochastic gradient descent~(DSGD) has been widely used for optimizing large-scale machine learning models, including both convex and non-convex models. With the rapid growth of model size, huge communication cost has been the bottleneck of traditional DSGD. Recently, many communication compression methods have been proposed. Memory-based distributed stochastic gradient descent~(M-DSGD) is one of the efficient methods since each worker communicates a sparse vector in each iteration so that the communication cost is small. Recent works propose the convergence rate of M-DSGD when it adopts vanilla SGD. However, there is still a lack of convergence theory for M-DSGD when it adopts momentum SGD. In this paper, we propose a universal convergence analysis for M-DSGD by introducing \emph{transformation equation}. The transformation equation describes the relation between traditional DSGD and M-DSGD so that we can transform M-DSGD to its corresponding DSGD. Hence we get the convergence rate of M-DSGD with momentum for both convex and non-convex problems. Furthermore, we combine M-DSGD and stagewise learning that the learning rate of M-DSGD in each stage is a constant and is decreased by stage, instead of iteration. Using the transformation equation, we propose the convergence rate of stagewise M-DSGD which bridges the gap between theory and practice.
\end{abstract}

\section{Introduction}
Many machine learning models can be formulated as the following empirical risk minimization problem:
\begin{align}\label{equation:object}
	\min_{\w\in \RB^d} F(\w) := \frac{1}{n}\sum_{i=1}^n f(\w;\zeta_i),
\end{align}
where $\w$ denotes the model parameter, $\zeta_i$ denotes the $i$th training data, $n$ is number of training data, $d$ is the size of models. SGD~\citep{Robbins&Monro:1951} is one of the efficient way to solve the empirical risk minimization problem. In each iteration, $\w$ is updated by $\w\leftarrow \w - \eta\nabla f(\w;\zeta_i)$. Comparing to the batch methods, like gradient descent, it only needs to calculate one gradient in each iteration.

With the rapid growth of data, using SGD to solve the empirical risk minimization problem is time-consuming. Hence, distributed stochastic gradient descent~(DSGD) has been the efficient method and many machine learning platforms~(e.g. TensorFlow, PyTorch) adopt it. With $p$ workers, it can be summarized as
\begin{align}\label{equation:dsgd}
	\w_{t+1} = \w_t - \eta_t\sum_{k=1}^p\g_{t,k}
\end{align}
where $\g_{t,k}$ is the update vector calculated by $k$th worker and usually satisfies unbiased estimation $\EB[\sum_{k=1}^p\g_{t,k}|\w_t] = \nabla F(\w_t)$. Workers parallel calculate $\g_{t,k}$ and the model parameter is updated by the summation of these $\g_{t,k}$ with learning rate $\eta_t$.

On the convergence of DSGD, it is equivalent to that of using single worker, which has the optimal rate $\mathcal{O}(1/\sqrt{T})$ for non-convex problems and $\mathcal{O}(1/T)$ for strongly convex problems~\citep{DBLP:journals/jmlr/DekelGSX12,DBLP:conf/icml/RakhlinSS12,DBLP:conf/kdd/LiZCS14}. Besides, communication is another important research area in the culture of distributed optimization. And recently, more and more large models, like DenseNet~\citep{DBLP:conf/cvpr/HuangLMW17}, Bert~\citep{DBLP:journals/corr/abs-1810-04805}, are used in machine learning. It leads to huge communication cost which cannot be ignored. Hence, communication compression has attracted much attention for further reducing training time.

One branch of this research area is low precision presentation~(also called quantization). On modern hardware, it uses $32$ bits to present a float number so that in DSGD, when one worker send or receive a $d$ dimension vector, the communication cost is $32d$. For a vector $\a \in \RB^d$, low precision presentation methods quantize $\a$ into $b$ bits presentation space, denoted as $Q(\a)$. It satisfies $\EB[Q(\a)|\a] = \a$ and the communication cost for $Q(\a)$ is $bd$. Usually they need to divide the $d$ coordinates into different buckets due to the quantization variance and then quantize them individually. Thus, the communication cost is $(bd+32s)$ and the compression ratio is $(bd+32s)/32d$, where $s$ is the number of buckets. It is easy to get that $(bd+32s)/32d \geq b/32 \geq 3\%$.


Another branch is sparse communication. For a vector $\a$, these methods make it sparse, denoted as $S(\a)\in \RB^d$ so that workers only need to send sparse vectors and can reduce the communication cost efficiently. In \citep{DBLP:conf/nips/WangSLCPW18,DBLP:conf/nips/WangniWLZ18}, they use stochastic sparsity technique to get $S(\a)$ with unbiased guarantee, i.e. $\EB[S(\a)] = \a$. Hence, these methods are equivalent to quantization ones mathematically~\citep{DBLP:conf/nips/WangSLCPW18}.  \citep{DBLP:conf/emnlp/AjiH17,DBLP:conf/iclr/LinHM0D18,DBLP:conf/nips/AlistarhH0KKR18,DBLP:conf/nips/StichCJ18} propose novel sparse communication methods that using memory gradient. Comparing to previous ones, $S(\a)$ is not necessarily the unbiased estimation of $\a$. It contains few coordinates of $\a$. After sending a sparse vector $S(\a)$ in each iteration, each worker stores those values which are not sent in the memory, i.e. $\a - S(\a)$. The $\a - S(\a)$ are called memory gradient and it will be used in the next iteration. These methods are called memory-based distributed stochastic gradient descent~(M-DSGD). \citep{DBLP:conf/emnlp/AjiH17,DBLP:conf/nips/AlistarhH0KKR18,DBLP:conf/nips/StichCJ18} are mainly based on vanilla SGD. ~\citep{DBLP:conf/nips/AlistarhH0KKR18} proves the convergence rate for convex problems and~\citep{DBLP:conf/nips/StichCJ18} proposes the convergence for both convex and non-convex problems. The convergence conditions of them are listed in Table~\ref{tab:convergence   condition}. ~\citep{DBLP:conf/iclr/LinHM0D18} adopts momentum SGD and get better performance. Empirical results on cifar10 and imagenet show that they only need to send a approximately $0.001d$ dimension vector in each iteration without loss of generalization, which means the compression ratio is smaller than $1\%$~\citep{DBLP:conf/iclr/LinHM0D18}. This is far better than that of quantization.  However, there is still a lack of convergence theory for M-DSGD when it adopts momentum SGD.

\begin{table}[!thb]
  \centering
  \caption{Convergence conditions in related works. }\label{tab:convergence condition}
  \begin{tabular}{|c|c|c|c|c|c|c||}
    \hline
    ~                                      & strong convex      & convex    & nonconvex           & momentum  \\ \hline
    \citep{DBLP:conf/nips/StichCJ18}       & $\surd$            & -         & -                   & no        \\ \hline
    \citep{DBLP:conf/nips/AlistarhH0KKR18} & $\surd$            & -         & $\surd$             & no        \\ \hline
    Ours                                   & $\surd$            & $\surd$   & $\surd$             & yes       \\
    \hline
  \end{tabular}
\end{table}

In this paper, we focus on the convergence rate of M-DSGD with momentum. The main results and contributions are summarized below:
\begin{itemize}
  \item We propose the \emph{transformation equation} for M-DSGD. It describes the relation between M-DSGD and traditional DSGD. According to the transformation equation, we can transform M-DSGD to its corresponding DSGD.
  \item When M-DSGD adopts $\beta$-momentum SGD, we prove the convergence rate for both convex and non-convex problems. When the momentum scalar $\beta$ is $0$, it degenerates to that of using vanilla SGD~\citep{DBLP:conf/emnlp/AjiH17} and we also get the convergence rate.
  \item We combine M-DSGD and stagewise learning~\citep{chen2018universal} that M-DSGD uses a constant learning rate in each stage, and decreases it by stage, which is usually adopted in practice. By the transformation equation, we prove the convergence rate of stagewise M-DSGD for a broad family of non-smooth and non-convex problems, which bridges the gap between theory and practice.
\end{itemize}

\section{Preliminary}
In this paper, we use $\|\cdot\|$ to denote $L_2$ norm, use $\w^*$ to denote the optimal solution of (\ref{equation:object}), use $\nabla f(\w; \IM_t)$ to denote one stochastic gradient with respect to mini-batch samples $\IM_t$ such that $\nabla f(\w;\mathcal{I}_t) = \frac{1}{|\IM_t|}\sum_{\zeta_i \in \IM_t} \nabla f(\w;\zeta_i)$ and $\EB_{\mathcal{I}_t}[\nabla f(\w;\mathcal{I}_t)|\w] = \nabla F(\w)$, use $\odot$ to denote dot product, use $\1$ to denote the vector $(1,1,\ldots,1)^T\in \RB^d$, use $\I$ to denote identity matrix. For a vector $\a$, we use $a^{(j)}$ to denote its $j$th coordinate value. We make the following definitions:
\begin{definition}\label{ass:bounded update vector}
	(bounded gradient)~$\g$ is the $G$-bounded~($G>0$) stochastic gradient of function $h(\cdot)$ if it satisfies $\EB[\g|\w] = \nabla h(\w)$, $\EB\|\g\|^2 \leq G^2, \forall \w$.
\end{definition}

\begin{definition}\label{ass:smooth loss function}
	(smooth function)~Function $h(\cdot)$ is $L$-smooth~($L>0$) if $\|\nabla h(\w) - \nabla h(\w')\| \leq L\|\w - \w'\|, \forall \w,\w'$, or equivalently $|h(\w) - h(\w') - \nabla h(\w')^T(\w - \w')| \leq \frac{L}{2}\|\w - \w'\|^2, \forall \w,\w'$.
\end{definition}

\begin{definition}\label{ass:strong convex object}
	(strong convex function) Function $h(\cdot)$ is $\mu$-strong convex~($\mu\geq 0$) if $h(\w) \geq h(\w')+\nabla h(\w')^T(\w - \w') + \frac{\mu}{2}\|\w - \w'\|^2, \forall \w,\w'$.
\end{definition}

\begin{definition}\label{ass:weak convex object}
	(weak convex function) Function $h(\cdot)$ is $c$-weak convex~($c\geq 0$) if $h(\w) \geq h(\w')+\nabla h(\w')^T(\w - \w') - \frac{c}{2}\|\w - \w'\|^2, \forall \w,\w'$.
\end{definition}

The first three definitions are common in both convex and non-convex optimization. Throughout this paper, we assume that $\EB_{\zeta_i}\|\nabla f(\w;\zeta_i)\|^2 \leq G^2, \forall \w$.

Recently, the weak convex property has attract much attention in non-convex optimization~\citep{DBLP:conf/icml/Allen-Zhu18,DBLP:conf/nips/Allen-Zhu18,chen2018universal}. For a $L$-smooth function, it must be $L$-weak convex. For a $c$-weak convex function, we can add one $L_2$ regularization to make it convex so that we can use convex optimization tools for a weak convex problems.

\section{Memory-based Distributed SGD}

\begin{algorithm}[t]
\caption{Memory-based Distributed SGD~(with momentum)}
\label{alg:mdsgd with momentum}
\begin{algorithmic}[1]
\STATE Initialization: $p$ workers, $\w_0$, $\beta \in [0,1)$, batch size $b$;
\STATE Set $\g_{-1,k} = \u_{0,k} = 0,k=1,\ldots,p,$
\FOR {$t=0,2,...T-1$}
\FOR {$k=1,2\ldots,p$, \textbf{each worker parallel}}
\STATE randomly picks one mini-batch training data $\IM_{t,k}$ with $|\IM_{t,k}| = b$;
\STATE Calculate the stochastic gradient $\frac{1}{b}\sum_{\zeta_i \in \IM_{t,k}} \nabla f(\w_t;\zeta_i)$;
\STATE $\g_{t,k} = \beta\g_{t-1,k} + \frac{1}{pb}\sum_{\zeta_i \in \IM_{t,k}} \nabla f(\w;\zeta_i)$;
\STATE Generate a sparse vector $\m_{t,k}\in \{0,1\}^d$;
\STATE Send $\m_{t,k}\odot(\g_{t,k}+\u_{t,k})$;
\STATE $\u_{t+1,k} = (\1 - \m_{t,k})\odot(\g_{t,k}+\u_{t,k})$, $k=1,2,\ldots,p$;
\ENDFOR
\STATE Aggregate: $\sum_{k=1}^p\m_{t,k}\odot(\g_{t,k}+\u_{t,k})$;
\STATE Update parameter: $\w_{t+1} = \w_t - \eta_t\sum_{k=1}^p\m_{t,k} \odot (\g_{t,k}+\u_{t,k})$;
\ENDFOR

\end{algorithmic}
\end{algorithm}

Assuming we have $p$ workers, the memory-based DSGD is presented in Algorithm~\ref{alg:mdsgd with momentum}. It can be implemented on many distributed platforms, like all-reduce, Parameter Server~\citep{DBLP:conf/osdi/LiAPSAJLSS14}. Data are divided into $p$ partitions and stored on $p$ workers. Each worker calculates update vector. After aggregating the update vectors $\m_{t,k}\odot(\g_{t,k}+\u_{t,k})$, it updates parameter $\w_t$. Since $\m_{t,k}$ is sparse, $\m_{t,k}\odot(\g_{t,k}+\u_{t,k})$ is sparse as well so that M-DSGD can reduce the communication cost. Besides, each worker will store those coordinates which are not be sent, denoted as $(\1 - \m_{t,k})\odot(\g_{t,k}+\u_{t,k})$. It is called memory gradient. In some related work~\citep{DBLP:conf/emnlp/AjiH17,DBLP:conf/nips/AlistarhH0KKR18}, it is called also residuals, accumulated error.

\subsection{Relation to Existing Sparse Communication Methods}
Assume we have got $\w_t, \u_{t,k},\g_{t-1,k}, k=1,2,\ldots,p$, the update rule of M-DSGD can be written as
\begin{align*}
  \g_{t,k} = & \beta\g_{t-1,k} + \frac{1}{pb}\sum_{\zeta_i \in \IM_{t,k}} \nabla f(\w_t;\zeta_i), \\
  \u_{t+1,k} = & (\1 - \m_{t,k})\odot(\g_{t,k}+\u_{t,k}), \\
  \w_{t+1} = & \w_t - \eta_t\sum_{k=1}^p\m_{t,k} \odot (\g_{t,k}+\u_{t,k}).
\end{align*}

The method in \citep{DBLP:conf/emnlp/AjiH17} is a special case of M-DSGD by setting $\beta = 0$, which means~\citep{DBLP:conf/emnlp/AjiH17} adopts the vanilla SGD.

\citep{DBLP:conf/nips/AlistarhH0KKR18} and~\citep{DBLP:conf/nips/StichCJ18} also use the memory gradient to make the communication sparse. Their update rules can be written as
      \begin{align*}
        \g_{t,k} = & \frac{1}{pb}\sum_{\zeta_i \in \IM_{t,k}} \nabla f(\w_t;\zeta_i), \nonumber \\
        \u_{t+1,k} = & (\1 - \m_{t,k})\odot(\eta_t\g_{t,k}+\u_{t,k}), \\
        \w_{t+1} = & \w_t - \sum_{k=1}^p\m_{t,k} \odot (\eta_t\g_{t,k}+\u_{t,k}).
      \end{align*}
      We can see that they also use the vanilla SGD. Compared to M-DSGD with $\beta = 0$, the difference is that their memory gradient $\u_{t+1,k}$ contains the learning rate $\eta_t$. By setting $\v_{t,k} = \frac{1}{\eta_t}\u_{t,k}$, we re-write the update rule for $\u_{t+1,k}$ and $\w_{t+1}$ as:
      \begin{align}
        \v_{t+1,k} = & \frac{\eta_t}{\eta_{t+1}}(\1 - \m_{t,k})\odot(\g_{t,k}+\v_{t,k}), \label{eq:contant eta 1} \\
        \w_{t+1} = & \w_t - \eta_t\sum_{k=1}^p\m_{t,k} \odot (\g_{t,k}+\v_{t,k}). \label{eq:contant eta 2}
      \end{align}
      We observe that the update rule for $\w_{t+1}$ is the same as that of M-DSGD. The difference is $\v_{t+1,k}$. In most convergence analysis for vanilla SGD, the learning rate $\eta_t$ is a constant or non-increasing. If $\eta_t$ is a constant, then it is totally the same as M-DSGD. If $\{\eta_t\}$ is a non-increasing sequence, on the one hand, in (\ref{eq:contant eta 1}), $\|\v_{t+1,k}\| \geq\| (\1 - \m_{t,k})\odot(\g_{t,k}+\v_{t,k})\|$. In the later convergence analysis, we will see that we should make the memory gradient norm $\|\v_{t+1,k}\|$ as small as possible. Hence the scalar $\eta_t/\eta_{t+1}$ is unnecessary and can be dropped. On the other hand, at the point of asynchronous updating view~\citep{DBLP:conf/iclr/LinHM0D18}, M-DSGD is more reasonable that $\u_{t+1,k}$ should not contain the learning rate $\eta_t$. Since the memory gradient denotes stale information, we should apply $\eta_{t+1}$, which is smaller than $\eta_t$, on $\u_{t+1,k}$ when we use it to get $\w_{t+2}$.

\citep{DBLP:conf/iclr/LinHM0D18} is the first work that adopts momentum SGD in M-DSGD. In \citep{DBLP:conf/iclr/LinHM0D18}, it uses a trick called momentum factor masking. Its update rule can be written as as
      \begin{align*}
        \hat{\g}_{t,k} = & \beta\g_{t-1,k} + \frac{1}{pb}\sum_{\zeta_i \in \IM_{t,k}}, \nabla f(\w;\zeta_i) \\
        \u_{t+1,k} = & (\1 - \m_{t,k})\odot(\hat{\g}_{t,k}+\u_{t,k}), \\
        \w_{t+1} = & \w_t - \eta_t\sum_{k=1}^p\m_{t,k} \odot (\hat{\g}_{t,k}+\u_{t,k}), \\
        \g_{t,k} = & (\1 - \m_{t,k})\odot \hat{\g}_{t,k}. ~\mbox{(momentum factor masking)}
      \end{align*}
      After getting $\u_{t+1,k}$, each worker applies the same $\m_{t,k}$ on $\hat{\g}_{t,k}$ to get $\g_{t,k}$.  \citep{DBLP:conf/iclr/LinHM0D18} considers the algorithm as a kind of asynchronous momentum SGD and the momentum factor masking can overcome the staleness effect. However, the $\m_{t,k}$ is designed mainly based on $\hat{\g}_{t,k}+\u_{t,k}$. It has nothing to do with $\hat{\g}_{t,k}$. The empirical results~\citep{DBLP:conf/iclr/LinHM0D18} on cifar10 using resnet110 show that the affect of momentum factor masking on top-1 accuracy is smaller than $1\%$.

\section{Transformation Equation}
For convenience, we define a diagonal matrix $\M_{t,k}\in \RB^{d\times d}$ such that $\mbox{diag}(\M_{t,k}) = \m_{t,k}$ to replace the symbol $\odot$. Then the update rule for $\w_t, \u_{t,k}$ can be written as
\begin{align}
	\w_{t+1} = & \w_t - \eta_t\sum_{k=1}^p\M_{t,k}(\g_{t,k}+\u_{t,k}), \label{eq:3}\\
	\u_{t+1,k} = & (\I - \M_{t,k})(\g_{t,k}+\u_{t,k}). \label{eq:4}
\end{align}
According to (\ref{eq:3}) and (\ref{eq:4}), we can  eliminate $\M_{t,k}$ and obtain
\begin{align}\label{eq:general update rule}
	\w_{t+1} - \eta_t\sum_{k=1}^p\u_{t+1,k} = \w_t - \eta_t\sum_{k=1}^p(\g_{t,k}+\u_{t,k}).
\end{align}
First, we consider the simplest cast that $\beta = 0$ to show the relation between traditional DSGD and M-DSGD. For convenience, we denote $\nabla f(\w_t;\IM_t) = \frac{1}{pb}\sum_{k=1}^p\sum_{\zeta_i \in \IM_{t,k}} \nabla f(\w_t;\zeta_i)$ which satisfies $\EB[\nabla f(\w_t;\IM_t)] = \nabla F(\w_t)$.

According to the above equation, we set $\z_t = \w_t - \eta_t\sum_{k=1}^p\u_{t,k}$ and obtain
\begin{align}
	\z_{t+1} = & \z_t - \eta_t\nabla f(\w_t;\IM_t) + (\eta_t - \eta_{t+1})\sum_{k=1}^p\u_{t+1,k} \nonumber \\
	         = & \underbrace{\z_t - \eta_t\nabla f(\z_t;\IM_t)}_{(\uppercase\expandafter{\romannumeral1})} + \underbrace{\eta_t(\nabla f(\z_t;\IM_t) - \nabla f(\w_t;\IM_t))}_{(\uppercase\expandafter{\romannumeral2})} + \underbrace{(\eta_t - \eta_{t+1})\sum_{k=1}^p\u_{t+1,k}}_{(\uppercase\expandafter{\romannumeral3})}\label{equation:transform}
\end{align}
According to equation (\ref{equation:transform}), we observe that:
\begin{itemize}
	\item for the term $(\uppercase\expandafter{\romannumeral1})$, it is the update rule for $\z_t$ in traditional DSGD;
	\item for the term $(\uppercase\expandafter{\romannumeral2})$, if $f(\w;\zeta_i)$ is smooth, we have $\|\eta_t(\nabla f(\z_t;\IM_t) - \nabla f(\w_t;\IM_t))\| \leq \mathcal{O}(\eta_t^2\|\sum_{k=1}^p\u_{t,k}\|);$
	\item for the term $(\uppercase\expandafter{\romannumeral3})$, if $|\eta_t-\eta_{t+1}|\leq \mathcal{O}(\eta_t^2)$, then $\|(\eta_t - \eta_{t+1})\sum_{k=1}^p\u_{t+1,k}\| \leq \mathcal{O}(\eta_t^2\|\sum_{k=1}^p\u_{t+1,k}\|).$
\end{itemize}
It implies that with certain assumptions, when we transform one traditional DSGD with initialization $\w_0$ and learning rate $\{\eta_t\}$ to M-DSGD, it is equivalent to adding one small noise scaled by $\eta_t^2$ in each iteration. To the best of our knowledge, for most DSGD's with convergence guarantee, the learning rate satisfies the condition $|\eta_t-\eta_{t+1}|\leq \mathcal{O}(\eta_t^2)$. For example, $\eta_t = \mathcal{O}(1/\sqrt{T})$ is a constant and $\eta_t = \mathcal{O}(1/t^\alpha), \alpha\in [0.5,1]$. When the noise is bounded and after being scaled by $\eta_t^2$, it will not affect the convergence of $\z_t$. What's more, when $\eta_t$ is small, $\w_t$ will get close to $\z_t$ which means $\w_t$ converges at the same time. Now we can conclude that both $\z_t$ and $\w_t$ converge to the optimal solution.

In fact, in equation (\ref{equation:transform}), we transform the update rule of $\w_t$ to that of $\z_t$ and get the convergence of $\w_t$ benefitting from the update rule of $\z_t$. For general $\g_{t,k}$, we have the following theorem:

\begin{theorem}\label{theorem:transformation}(Transformation)
Let $\{\w_t\}$ be a sequence generated by Algorithm \ref{alg:mdsgd with momentum} with learning rate $\{\eta_t\}$. We set $\z_t$ to be some \emph{linear combination} of $\w_i,\u_{i,k},\g_{i,k}, i=0,\ldots,t, k=1,\ldots,p$, and define $\gamma_t = \psi_t(\eta_0, \eta_1,\ldots,\eta_t)$, where $\psi_t(\cdot)$ is some function. Assume the following condition holds on:
\begin{align}
  & \EB\|\z_t - \w_t\|^2 \leq A\gamma_t^2, \forall t\geq 0; \label{eq:close} \\
  & \z_{t+1} = \z_t - \gamma_t \d_t + \alpha_t\e_t, \label{eq:general transformation equation}
\end{align}
where $|\alpha_t| \leq \delta\gamma_t^2\leq Q$, $\EB[\d_t|\w_t] = \nabla F(\w_t)$, $\EB\|\d_t\|^2\leq D^2, \EB\|\e_t\|^2\leq E^2, \forall t$. Then we call (\ref{eq:general transformation equation}) the \textbf{transformation equation}. If $F(\cdot)$ is $L$-smooth with $G$-bounded stochastic gradient, we have
\begin{align*}
  \sum_{t=0}^{T-1}\gamma_t\EB[\|\nabla F(\w_t)\|^2|\w_t] \leq F(\w_0) - F(\w^*) + C\sum_{t=0}^{T-1}\gamma_t^2.
\end{align*}
where $C = LG\sqrt{A} + GE\delta + L(D^2 + E^2Q\delta)$.
\end{theorem}

In Theorem \ref{theorem:transformation}, we only need $\z_t$ to be the \emph{linear combination} of $\w_i,\u_{i,k},\g_{i,k}, i=0,\ldots,t, k=1,\ldots,p$ so that it is easy to conduct such a $\z_t$. Although $\d_t$ is an unbiased estimation of full gradient at $\w_t$, benefitting from (\ref{eq:close}) which implies $\z_t$ and $\w_t$ are close enough and $\|\alpha_t\e_t\|$ is the same order of magnitude as the  variance of $\gamma_t\d_t$,  (\ref{eq:general transformation equation}) can be seen as updating $\z_t$ by DSGD with learning rate $\gamma_t$. Hence, (\ref{eq:general transformation equation}) transform the update rule of $\w_t$ to that of $\z_t$ and we call it the transformation equation. It describes the relation between M-DSGD and traditional DSGD. If
\begin{align}\label{equation:learning rate condition}
  \sum_{t=0}^{T-1}\gamma_t \rightarrow \infty, \sum_{t=0}^{T-1}\gamma_t^2/\sum_{t=0}^{T-1}\gamma_t \rightarrow 0, \mbox{as } T\rightarrow \infty,
\end{align}
then we can randomly choose $\w$ from $\{\w_0,\w_1,\ldots,\w_{T-1}\}$ with probability $P(\w = \w_t) = \gamma_t/\sum_{t=0}^{T-1}\gamma_t$, and get that $\EB\|\nabla F(\w)\|^2 \rightarrow 0$.

\section{Convergence}
In this section, we are going to prove the convergence of $\{\w_t\}$ of M-DSGD with $\beta$-momentum for both convex and non-convex problems. For convenience, we denote $\tilde{\u}_t = \sum_{k=1}^p \u_{t,k}$ and $\tilde{\g}_t = \sum_{k=1}^p \g_{t,k}$. Then according to (\ref{eq:general update rule}), we have the update rule for $\w_t$:
\begin{align}
	\w_{t+1} - \eta_t\tilde{\u}_{t+1} = \w_t - \eta_t(\tilde{\g}_t+\tilde{\u}_t),
\end{align}
where $\tilde{\g}_t = \beta\tilde{\g}_{t-1} + \nabla f(\w_t;\mathcal{I}_t)$.

According to Theorem \ref{theorem:transformation}, our main task is establishing the transformation equation.

\begin{lemma}\label{lemma:local msgd}
Let $\tilde{\g}_t = \nabla f(\w_t;\mathcal{I}_t) + \beta\tilde{\g}_{t-1}, \beta\in [0,1)$. By setting
\begin{align}
  \z_t = \w_t + \rho_{t-1}\tilde{\g}_{t-1} - \eta_t\tilde{\u}_t,
\end{align}
where $\beta\rho_t = \beta\eta_t+\rho_{t-1}$, we have
\begin{align}\label{equation:transform with local momentum}
  \z_{t+1} =  \z_t -(\eta_t - \rho_t)\nabla f(\w_t;\IM_t) + (\eta_t - \eta_{t+1})\tilde{\u}_{t+1}.
\end{align}
\end{lemma}

Lemma \ref{lemma:local msgd} gives out the transformation equation of M-DSGD with $\beta$-momentum. We can see that $\z_t$ is a linear combination of $\w_t, \tilde{\g}_{t-1}$ and $\tilde{\u}_t$. Since $\rho_t = \eta_t + \frac{1}{\beta}\rho_{t-1}$ and $0\leq \beta <1$, it is easy to make $\rho_t\rightarrow \infty$. We should design $\eta_t$ carefully. We propose two strategies:
\begin{itemize}
  \item $\eta_t = \eta/\sqrt{T}$ is a small constant, then $\rho_t = \beta\eta/((\beta-1)\sqrt{T})<0$, and $\eta_t - \rho_t = \eta/((1-\beta)\sqrt{T})$;
  \item $\eta_t = \eta(1/t^\alpha - \beta/(t+1)^\alpha), \alpha\in[0.5,1]$, then $\rho_t = -\eta/(t+1)^\alpha<0$. and $\eta_t - \rho_t = \eta/(\beta t^\alpha)$.
\end{itemize}
It is easy to verify that both of the two strategies satisfy (\ref{eq:close}) and the learning rate condition~(\ref{equation:learning rate condition}). Specifically, we have the following lemma:
\begin{lemma}\label{lemma:bounded g}
  Assume $F(\cdot)$ has the $G$-bounded stochastic gradient, and $\EB\|\tilde{\u}_{t}\|^2\leq U^2, \forall t$. $\{\z_t\}$ is defined in Lemma \ref{lemma:local msgd}. Then we have
  \begin{align*}
    \EB\|\tilde{\g}_t\|^2 \leq \frac{G^2}{(1-\beta)^2}
  \end{align*}
  and if $\eta_t, \rho_t$ is defined as one of the above two strategies, we have
  \begin{align*}
    \EB\|\z_t - \w_t\|^2 \leq \frac{2G^2}{(1-\beta)^2}\rho_{t-1}^2 + 2U^2\eta_t^2 \leq \mathcal{O}(\rho_{t-1}^2)
  \end{align*}
\end{lemma}

Then we get the following convergence rate of M-DSGD:

\begin{theorem}\label{theorem:local msgd convex}
	(strong convex case)~Let $\tilde{\g}_t = \nabla f(\w;\mathcal{I}_t) + \beta\tilde{\g}_{t-1}, \beta\in [0,1)$ and $\z_t$ is defined in Lemma \ref{lemma:local msgd}.  Assume $F(\cdot)$ is $L$-smooth, $\mu$-strong convex with $G$-bounded stochastic gradient, and $\EB\|\tilde{\u}_{t}\|^2\leq U^2, \EB\|\z_t - \w^*\|^2\leq B^2,\forall t$. By setting $\rho_t = -\beta/(\mu (t+1)), \eta_t = (1 - (\beta t)/(t+1))/(\mu t), t\geq 1$, we have
  \begin{align*}
    \frac{1}{\lceil T/2 \rceil}\sum_{t=T-\lceil T/2 \rceil}^{T-1}\EB(F(\w_t) - F(\w^*))\leq & \frac{3C + 2G \sqrt{\frac{2G^2\beta^2}{(1-\beta)^2} + 2U^2}}{\mu T},
  \end{align*}
  where $C = \max\{4G^2, 2LB\sqrt{2G^2\beta^2/(1-\beta)^2 + 2U^2} + 2\mu UB + 2G^2 + 2U^2\}$.
\end{theorem}

\begin{theorem}\label{theorem:local msgd general convex}
	(convex case)~Let $\tilde{\g}_t = \nabla f(\w;\mathcal{I}_t) + \beta\tilde{\g}_{t-1}, \beta\in [0,1)$ and $\z_t$ is defined in Lemma \ref{lemma:local msgd}.  Assume $F(\cdot)$ is convex with $G$-bounded stochastic gradient, and $\EB\|\tilde{\u}_{t}\|^2\leq U^2, \EB\|\z_t - \w^*\|^2\leq B^2,\forall t$. By setting $\rho_t = -\beta/\sqrt{t+2}, \eta_t = (1 - \beta \sqrt{t+1}/\sqrt{t+2})/\sqrt{t+1}$, we have
	\begin{align*}
        \sum_{t=0}^{T-1}\frac{2}{\sqrt{t+1}}\EB(F(\w_t) - F(\w^*)) \leq \|\w_{0} - \w^*\|^2 + \sum_{t=0}^{T-1}\frac{C}{t+1},
  \end{align*}
    where $C = 2G\sqrt{2G^2\beta^2/(1-\beta)^2 + 2U^2} + 2UB + 2G^2 + 2U^2$. It implies the $\mathcal{O}(\log(T)/\sqrt{T})$ convergence rate.
\end{theorem}

\begin{theorem}\label{theorem:local msgd nonconvex}
	(non-cnovex case)~Let $\tilde{\g}_t = \nabla f(\w;\mathcal{I}_t) + \beta\tilde{\g}_{t-1}, \beta\in [0,1)$ and $\z_t$ is defined in Lemma \ref{lemma:local msgd}. Assume $F(\cdot)$ is $L$-smooth with $G$-bounded stochastic gradient and $\EB\|\tilde{\u}_{t}\|^2\leq U^2$. By setting $\eta_t = \eta, \rho_t = (\beta\eta)/(\beta-1)$, we have
	\begin{align*}
        \frac{1}{(1-\beta)T}\sum_{t=0}^{T-1}\|\nabla F(\w_t)\|^2 \leq \frac{F(\w_0) - F(\w^*)}{T\eta} + C\eta,
    \end{align*}
    where $C = LG^2\beta/(1-\beta)^3 + LGU/(1-\beta) + LG^2/(2(1-\beta)^2)$. By taking $\eta = \mathcal{O}(1/\sqrt{T})$, it is easy to get $\mathcal{O}(1/\sqrt{T})$ convergence rate.
\end{theorem}

If we design $\g_{t,k}$ as stochastic batch gradient of $F(\w)$, which means $\tilde{\g}_t = \nabla f(\w_t;\mathcal{I}_t)$, it is a special case of M-DSGD with momentum by setting $\beta=0$. For the strong convex and smooth case, by setting $\eta_t = 1/\mu t$, we get the $\mathcal{O}(1/T)$ convergence rate. For the general convex case, by setting $\eta_t = 1/\sqrt{t+1}$, we get the $\mathcal{O}(\log(T)/\sqrt{T})$ convergence rate. For non-convex and smooth case, by setting $\eta_t = \eta = \mathcal{O}(1/\sqrt{T})$, we get the $\mathcal{O}(1/\sqrt{T})$ convergence rate. Please note that (\ref{eq:contant eta 1})(\ref{eq:contant eta 2}) can also be transformed to the such a formulation that $\z_{t+1} = \z_t - \eta_t\nabla f(\w_t; \IM_t)$, where $\z_t = \w_t - \eta_t\sum_{k=1}^{p}\v_{t,k}$. So it has the same convergence rate.

\section{M-DSGD meets stagewise learning}
In the previous analysis for convergence of M-DSGD, we need to set the learning rate $\eta_t$ be a constant or $\mathcal{O}(1/t^{\alpha})$. This is far from the practice. In fact, we never set a constant learning rate when training models. Usually, we decrease the learning rate after passing through the training data several times. For example, \citep{DBLP:conf/cvpr/HeZRS16} decreases the learning rate by $\eta \leftarrow 0.1\eta$ at 32k and 48k iterations when training resnet on imagenet. What's more, many models include non-smooth operations, like ReLu, max pooling. This is also far from the convergence condition in previous theorems. Recently, \citep{chen2018universal} propose the stagewise learning to bridge the gap between theory and practice. In this section, we use stagewise learning for M-DSGD.

For convenience, we denote Algorithm~\ref{alg:mdsgd with momentum} with constant learning rate $\eta$ as $\AM(\phi(\cdot),\tilde{\w},\eta,\beta, T)$. $\phi(\cdot)$ is the function to optimize, $\tilde{\w}$ is initialization, $\eta$ is a constant learning rate, $\beta$ is the momentum scalar, $T$ is the iteration numbers. Then we have
\begin{lemma}\label{lemma:onestage mdgsd}
  Let $\tilde{\w}^{+} = \AM(\phi(\cdot),\tilde{\w},\eta,\beta, T)$. We define $\{\w_t\}, \{\tilde{\u}_t\}$ to be the sequence produced by $\AM(\phi(\cdot),\tilde{\w},\eta,T)$ so that $\tilde{\w} = \w_0, \w_{t+1} - \eta_t\tilde{\u}_{t+1} = \w_t - \eta_t(\tilde{\g}_t+\tilde{\u}_t)$, and $\tilde{\w}^{+} =\sum_{t=0}^{T-1}\w_t/T$. Let $\{\z_t\}$ be the sequence transformed by Lemma \ref{lemma:local msgd}. Assume $\EB\|\tilde{\u}_t\|^2\leq U^2$, $\phi(\cdot)$ is convex with $G$-bounded stochastic gradient, we have
  \begin{align*}
    \EB\phi(\tilde{\w}^{+}) - \phi(\w_\phi^*) \leq \frac{1-\beta}{2\eta T}\|\w - \w_\phi^*\|^2 + C\eta
  \end{align*}
  where $C = G\sqrt{2G^2\beta^2/(1-\beta)^4 + 2U^2} + G^2/(2-2\beta)$, $\w_{\phi}^* = \mathop{\arg\min}_\w \phi(\w)$.
\end{lemma}
Here we set $\phi(\cdot)$ to be convex but not necessarily strong convex.

Lemma \ref{lemma:onestage mdgsd} implies that M-DSGD with constant learning rate satisfies the condition of stagewise learning~(Theorem 1 in~\citep{chen2018universal}). And constant learning rate makes the transformation equation simple. Thus, we can use stagewise learning. We define
\begin{align}
  \tilde{\w}_{s+1} = \AM(F_{s,\gamma}(\cdot), \tilde{\w}_s, \eta_s, \beta, T_s)
\end{align}
where
\begin{align}
  F_{s,\gamma}(\w) = F(\w) + \frac{1}{2\gamma}\|\w - \tilde{\w}_s\|^2
\end{align}

Let $\{\w_{s,t}\}, \{\tilde{\u}_{s,t}\}$ be the sequence produced by $\AM(F_{s,\gamma}(\cdot), \tilde{\w}_s, \eta_s, \beta, T_s)$, which means $\w_{s,t+1} - \eta_s\tilde{\u}_{s,t+1} = \w_{s,t} - \eta_s(\tilde{\g}_{s,t}+\tilde{\u}_{s,t})$. If $F(\cdot)$ is $c$-weak convex, then $F_{s,\gamma}$ is $(\gamma^{-1} - c)$-strong convex when $\gamma < 1/c$. Hence, we can apply Lemma \ref{lemma:onestage mdgsd} on $F_{s,\gamma}(\w)$. Specifically, we have the following result:

\begin{theorem}\label{theorem:stagewise mdsgd}
  Assume $F(\cdot)$ is $c$-weak convex with $G$-bounded stochastic gradient and $\EB\|\w_{s,t} - \w^*\|^2 \leq B^2, \EB\|\tilde{\u}_{s,t}\|^2 \leq U^2, \EB F(\tilde{\w}_s) \leq F, \forall s,t$.  By setting $\gamma = 1/(2c)$, $\eta_s = \eta_0/(s+1), \eta_sT_s\geq 12\gamma$, we have
  \begin{align*}
    \frac{(1+\beta)\gamma}{4S(S+1)} \sum_{s=0}^{S-1}(s+1)\EB\|\nabla F_{\gamma}(\tilde{\w}_{s})\|^2 \leq \frac{F-F(\w^*)+3\hat{C}\eta_0}{S+1}
  \end{align*}
  where $F_{\gamma}(\cdot)$ is defined as $$F_{\gamma}(\w) = \min_{\w'}F(\w') + \frac{1}{2\gamma}\|\w - \w'\|^2$$, and $\hat{C} = \hat{G}\sqrt{2\hat{G}^2\beta^2/(1-\beta)^4 + 2U^2} + \hat{G}^2/(2-2\beta), \hat{G} = \sqrt{2G^2 + 4B^2/\gamma^2}$.
\end{theorem}

In both Lemma \ref{lemma:onestage mdgsd} and Theorem \ref{theorem:stagewise mdsgd}, we do not need the smooth assumption for $f(\w;\zeta_i)$ or $F(\w)$. Hence, stagewise M-DSGD can solve a broad family of non-smooth and non-convex problems.

\section{Choice of $\m_{t,k}$}
In the convergence theorems, we need $\EB\|\tilde{\u}_t\|^2$ to be bounded. Since $\EB\|\tilde{\u}_t\|^2 = \EB\|\sum_{k=1}^{p}\u_{t,k}\|^2 \leq p\sum_{k=1}^{p}\EB\|\u_{t,k}\|^2$ to be bounded, we only need $\EB\|\u_{t,k}\|^2$ to be bounded. According to the update rule for $\u_{t,k}$:
\begin{align*}
  \g_{t,k} = & \beta\g_{t-1,k} + \frac{1}{pb}\sum_{\zeta_i \in \IM_{t,k}} \nabla f(\w_t;\zeta_i) \\
  \u_{t+1,k} = & (\1 - \m_{t,k})\odot(\g_{t,k}+\u_{t,k})
\end{align*}
and recent works~\citep{DBLP:conf/emnlp/AjiH17,DBLP:conf/nips/AlistarhH0KKR18,DBLP:conf/nips/StichCJ18,DBLP:conf/iclr/LinHM0D18} propose three strategies for generating $\m_{t,k}$: random strategy, top-K strategy and approximate top-K strategy, if $\g_{t,k}$ is bounded, $\u_{t,k}$ is bounded as well under the three strategies. Specifically, we have the following lemma:
\begin{lemma}\label{lemma:top-k}
  If $\m_{t,k}\in \{0,1\}^d$ adopts random strategy or top-K strategy with $\|\m_{t,k}\|_0 = q,\forall t,k$, then we have $$\EB\|\tilde{\u}_{t}\|^2 \leq \frac{2(d-q)(2d+q)G^2}{(1-\beta)^2q^2}.$$
\end{lemma}

It implies that $\|\tilde{\u}_{t}\|$ is nothing to do with the number of workers $p$. According to the convergence theorems, we should make $\|\tilde{\u}_{t}\|$ as small as possible and usually the top-K strategy is the best one~\citep{DBLP:conf/nips/StichCJ18}.
\section{Conclusion}
In this paper, we propose the transformation equation for theoretical analysis of M-DSGD and get the convergence rate of M-DSGD with momentum. Transformation equation describes the relation between traditional DSGD and M-DSGD. By transformation equation, we find that M-DSGD can be seen as adding one small noise on DSGD which can not affect the convergence. Thus, we get the convergence rate of M-DSGD with momentum easily for both convex and non-convex optimization. We combine M-DSGD and stagewise learning that the learning rate of M-DSGD in each stage is a constant and is decreased by stage, which is more practical. We propose the convergence rate of stagewise M-DGSD by transformation equation for a broad family of non-smooth, non-convex problems, which bridges the gap between theory and practice.

\bibliography{ref}

\end{document}

%% file: defs.tex
\def\a{{\bf a}}

\def\d{{\bf d}}
\def\e{{\bf e}}

\def\g{{\bf g}}

\def\m{{\bf m}}

\def\u{{\bf u}}
\def\v{{\bf v}}
\def\w{{\bf w}}

\def\z{{\bf z}}

\def\I{{\bf I}}

\def\M{{\bf M}}

\def\0{{\bf 0}}
\def\1{{\bf 1}}
\def\2{{\bf 2}}
\def\3{{\bf 3}}
\def\4{{\bf 4}}
\def\5{{\bf 5}}
\def\6{{\bf 6}}
\def\7{{\bf 7}}
\def\8{{\bf 8}}
\def\9{{\bf 9}}

\def\AM{{\mathcal A}}

\def\IM{{\mathcal I}}

\def\EB{{\mathbb E}}

\def\RB{{\mathbb R}}

%% file: con_mdsgd.bbl
\begin{thebibliography}{17}
\providecommand{\natexlab}[1]{#1}
\providecommand{\url}[1]{\texttt{#1}}
\expandafter\ifx\csname urlstyle\endcsname\relax
  \providecommand{\doi}[1]{doi: #1}\else
  \providecommand{\doi}{doi: \begingroup \urlstyle{rm}\Url}\fi

\bibitem[Aji and Heafield(2017)]{DBLP:conf/emnlp/AjiH17}
Alham~Fikri Aji and Kenneth Heafield.
\newblock Sparse communication for distributed gradient descent.
\newblock In \emph{Proceedings of the 2017 Conference on Empirical Methods in
  Natural Language Processing}, pages 440--445, 2017.

\bibitem[Alistarh et~al.(2018)Alistarh, Hoefler, Johansson, Konstantinov,
  Khirirat, and Renggli]{DBLP:conf/nips/AlistarhH0KKR18}
Dan Alistarh, Torsten Hoefler, Mikael Johansson, Nikola Konstantinov, Sarit
  Khirirat, and C{\'{e}}dric Renggli.
\newblock The convergence of sparsified gradient methods.
\newblock In \emph{Advances in Neural Information Processing Systems}, pages
  5977--5987, 2018.

\bibitem[Allen{-}Zhu(2018{\natexlab{a}})]{DBLP:conf/icml/Allen-Zhu18}
Zeyuan Allen{-}Zhu.
\newblock Katyusha {X:} practical momentum method for stochastic
  sum-of-nonconvex optimization.
\newblock In \emph{Proceedings of the 35th International Conference on Machine
  Learning}, pages 179--185, 2018{\natexlab{a}}.

\bibitem[Allen{-}Zhu(2018{\natexlab{b}})]{DBLP:conf/nips/Allen-Zhu18}
Zeyuan Allen{-}Zhu.
\newblock How to make the gradients small stochastically: Even faster convex
  and nonconvex {SGD}.
\newblock In \emph{Advances in Neural Information Processing Systems}, pages
  1165--1175, 2018{\natexlab{b}}.

\bibitem[Chen et~al.(2019)Chen, Yuan, Yi, Zhou, Chen, and
  Yang]{chen2018universal}
Zaiyi Chen, Zhuoning Yuan, Jinfeng Yi, Bowen Zhou, Enhong Chen, and Tianbao
  Yang.
\newblock Universal stagewise learning for non-convex problems with convergence
  on averaged solutions.
\newblock In \emph{International Conference on Learning Representations}, 2019.

\bibitem[Dekel et~al.(2012)Dekel, Gilad{-}Bachrach, Shamir, and
  Xiao]{DBLP:journals/jmlr/DekelGSX12}
Ofer Dekel, Ran Gilad{-}Bachrach, Ohad Shamir, and Lin Xiao.
\newblock Optimal distributed online prediction using mini-batches.
\newblock \emph{Journal of Machine Learning Research}, 13:\penalty0 165--202,
  2012.

\bibitem[Devlin et~al.(2018)Devlin, Chang, Lee, and
  Toutanova]{DBLP:journals/corr/abs-1810-04805}
Jacob Devlin, Ming{-}Wei Chang, Kenton Lee, and Kristina Toutanova.
\newblock {BERT:} pre-training of deep bidirectional transformers for language
  understanding.
\newblock \emph{CoRR}, abs/1810.04805, 2018.

\bibitem[He et~al.(2016)He, Zhang, Ren, and Sun]{DBLP:conf/cvpr/HeZRS16}
Kaiming He, Xiangyu Zhang, Shaoqing Ren, and Jian Sun.
\newblock Deep residual learning for image recognition.
\newblock In \emph{2016 {IEEE} Conference on Computer Vision and Pattern
  Recognition}, pages 770--778, 2016.

\bibitem[Huang et~al.(2017)Huang, Liu, van~der Maaten, and
  Weinberger]{DBLP:conf/cvpr/HuangLMW17}
Gao Huang, Zhuang Liu, Laurens van~der Maaten, and Kilian~Q. Weinberger.
\newblock Densely connected convolutional networks.
\newblock In \emph{2017 {IEEE} Conference on Computer Vision and Pattern
  Recognition, {CVPR} 2017, Honolulu, HI, USA, July 21-26, 2017}, pages
  2261--2269, 2017.

\bibitem[Li et~al.(2014{\natexlab{a}})Li, Andersen, Park, Smola, Ahmed,
  Josifovski, Long, Shekita, and Su]{DBLP:conf/osdi/LiAPSAJLSS14}
Mu~Li, David~G. Andersen, Jun~Woo Park, Alexander~J. Smola, Amr Ahmed, Vanja
  Josifovski, James Long, Eugene~J. Shekita, and Bor{-}Yiing Su.
\newblock Scaling distributed machine learning with the parameter server.
\newblock In \emph{11th {USENIX} Symposium on Operating Systems Design and
  Implementation}, pages 583--598, 2014{\natexlab{a}}.

\bibitem[Li et~al.(2014{\natexlab{b}})Li, Zhang, Chen, and
  Smola]{DBLP:conf/kdd/LiZCS14}
Mu~Li, Tong Zhang, Yuqiang Chen, and Alexander~J. Smola.
\newblock Efficient mini-batch training for stochastic optimization.
\newblock In \emph{The 20th {ACM} {SIGKDD} International Conference on
  Knowledge Discovery and Data Mining}, pages 661--670, 2014{\natexlab{b}}.

\bibitem[Lin et~al.(2018)Lin, Han, Mao, Wang, and
  Dally]{DBLP:conf/iclr/LinHM0D18}
Yujun Lin, Song Han, Huizi Mao, Yu~Wang, and Bill Dally.
\newblock Deep gradient compression: Reducing the communication bandwidth for
  distributed training.
\newblock In \emph{6th International Conference on Learning Representations},
  2018.

\bibitem[Rakhlin et~al.(2012)Rakhlin, Shamir, and
  Sridharan]{DBLP:conf/icml/RakhlinSS12}
Alexander Rakhlin, Ohad Shamir, and Karthik Sridharan.
\newblock Making gradient descent optimal for strongly convex stochastic
  optimization.
\newblock In \emph{Proceedings of the 29th International Conference on Machine
  Learning}, 2012.

\bibitem[Robbins and Monro(1951)]{Robbins&Monro:1951}
H.~Robbins and S.~Monro.
\newblock A stochastic approximation method.
\newblock \emph{Annals of Mathematical Statistics}, 22:\penalty0 400--407,
  1951.

\bibitem[Stich et~al.(2018)Stich, Cordonnier, and
  Jaggi]{DBLP:conf/nips/StichCJ18}
Sebastian~U. Stich, Jean{-}Baptiste Cordonnier, and Martin Jaggi.
\newblock Sparsified {SGD} with memory.
\newblock In \emph{Advances in Neural Information Processing Systems}, pages
  4452--4463, 2018.

\bibitem[Wang et~al.(2018)Wang, Sievert, Liu, Charles, Papailiopoulos, and
  Wright]{DBLP:conf/nips/WangSLCPW18}
Hongyi Wang, Scott Sievert, Shengchao Liu, Zachary~B. Charles, Dimitris~S.
  Papailiopoulos, and Stephen Wright.
\newblock {ATOMO:} communication-efficient learning via atomic sparsification.
\newblock In \emph{Advances in Neural Information Processing Systems}, pages
  9872--9883, 2018.

\bibitem[Wangni et~al.(2018)Wangni, Wang, Liu, and
  Zhang]{DBLP:conf/nips/WangniWLZ18}
Jianqiao Wangni, Jialei Wang, Ji~Liu, and Tong Zhang.
\newblock Gradient sparsification for communication-efficient distributed
  optimization.
\newblock In \emph{Advances in Neural Information Processing Systems}, pages
  1306--1316, 2018.

\end{thebibliography}
